\documentclass[11pt,technote,onecolumn]{IEEEtran}
\usepackage{times}
\usepackage{epsfig}
\usepackage{graphicx}
\usepackage{amsmath}
\usepackage{amssymb}
\usepackage{bm}
\usepackage{multirow}
\usepackage{color}
\usepackage[noend]{algpseudocode}
\usepackage{algorithmicx,algorithm}
\usepackage{array}
\usepackage{CJK}
\usepackage{subfigure} 
\ifCLASSINFOpdf
\else
\fi
\begin{document}
%
\title{Scalable Deep Compressive Sensing}
%
%

\author{Zhonghao Zhang, Yipeng Liu ~\IEEEmembership{Senior Member,~IEEE}, Xingyu Cao, Fei Wen, Ce Zhu ~\IEEEmembership{Fellow,~IEEE}
	\thanks{This research is supported in part by the National Natural Science Foundation of China (NSFC, No. 62020106011, No. U19A2052), in part by Sichuan Science and Technology Program (No. 2019YFH008). The corresponding author is Yipeng Liu.}
	\thanks{Zhonghao Zhang, Yipeng Liu, Xingyu Cao and Ce Zhu are with School of Information and Communication Engineering, University of Electronic Science and Technology of China (UESTC), Chengdu 611731, China. (email:yipengliu@uestc.edu.cn).}
		\thanks{Fei Wen is with the Department of Electronic Engineering, Shanghai Jiao Tong University, Shanghai 200240, China.}
	}

%
%

\markboth{Paper Name,~Vol.~XXX, No.~XXX, MOnth~Year}%
{Shell \MakeLowercase{\textit{et al.}}: Bare Demo of IEEEtran.cls for IEEE Journals}
%



\maketitle

\begin{abstract}
Deep learning has been used to image compressive sensing (CS) for enhanced reconstruction performance. However, most existing deep learning methods train different models for different subsampling ratios, which brings additional hardware burden. In this paper, we develop a general framework named scalable deep compressive sensing (SDCS) for the scalable sampling and reconstruction (SSR) of all existing end-to-end-trained models. In the proposed way, images are measured and initialized linearly. Two sampling masks are introduced to flexibly control the subsampling ratios used in sampling and reconstruction, respectively. To make the reconstruction model adapt to any subsampling ratio, a training strategy dubbed scalable training is developed. In scalable training, the model is trained with the sampling matrix and the initialization matrix at various subsampling ratios by integrating different sampling matrix masks. Experimental results show that models with SDCS can achieve SSR without changing their structure while maintaining good performance, and SDCS outperforms other SSR methods.
\end{abstract}


%
\IEEEpeerreviewmaketitle

\section{Introduction}
\IEEEPARstart{C}{ompressive} sensing (CS) is a technique that simultaneously samples and compresses signals.
And the signal is sampled and reconstructed at a ratio which can be much lower than the Nyquist rate. CS has been applied in a series of applications, such as single pixel imaging (SPI)~\cite{duarte2008single}, magnetic resonance imaging (MRI)~\cite{liu2017hybrid} and wireless broadcast~\cite{Yin2016Compressive}.

The sampling process of CS can be expressed as $\mathbf{y} = \mathbf{Ax} $
, where $\mathbf{x} \in {\mathbb{R}^N}$ is the original signal, $\mathbf{y} \in  {\mathbb{R}^M}$ denotes the measurement, $\mathbf{A}\in {\mathbb{R}^{M \times N}}$ is the sampling matrix with $ M < N $ and $M/N$ is the CS ratio.
The signal recovery from $\mathbf{y}$ is under-determined, and it is usually be carried out by solving an optimization problem as follows:
\begin{align}
\label{CS_problem}
\mathop {\min }\bm{\limits_\mathbf{x}} \mathfrak{R}(\mathbf{x})\,{\text{,~~s}}{\text{.~t}}{\text{.~~}}\mathbf{y} = \mathbf{A x},
\end{align}
where  $\mathfrak{R}(\mathbf{x})$ is the regularization term. In this paper, we mainly focus on the visual image CS~\cite{kulkarni2016reconnet} which has been applied in SPI~\cite{duarte2008single,kulkarni2016reconnet} and wireless broadcast~\cite{Yin2016Compressive,li2013new}. And since block-by-block sampling and reconstruction~\cite{dong2014compressive,Dinh2017ietative,kulkarni2016reconnet,zhang2018ista,2020AMP} would bring less burden to the hardware, we mainly focus on the block-based visual image CS problem.

To solve the problem (\ref{CS_problem}), model-based methods introduce various hand-crafted regularizers to represent the prior information of images, such as sparsity~\cite{mallat1999wavelet,elad2010sparse}, low rank~\cite{dong2014compressive,liu2019low} and so on~\cite{metzler2016denoising, Wu2019deep}. And many non-linear iterative algorithms can be applied for image reconstruction, such as fast iterative shrinkage-thresholding algorithm (FISTA)~\cite{beck2009fast}, approximate message passing (AMP)~\cite{donoho2009message}, etc. These methods~\cite{dong2014compressive,li2020scalable} usually have theoretical guarantees and work well using sampling matrices with different CS ratios. However, their performance needs to be further improved.

In recent years, deep learning has achieved great success in image inverse problems~\cite{rick2017one,dong2018denoising}. Among them, models for image CS can be cast into two categories: traditional deep learning models and deep unfolding models.
Traditional deep learning models are stacked by non-linear computational layers, such as autoencoder~\cite{mousavi2015deep}, convolutional neural network (CNN)~\cite{kulkarni2016reconnet,shi2019image,shi2020video} and generative adversarial network (GAN)~\cite{pmlr-v70-bora17a}. 
Moreover, some techniques can be applied for better performance, such as the residual connection~\cite{he2016deep} for better training. These models map the measurement to the output without considering the prior information of images. 
Although they can reconstruct high-quality images with a high speed, there is no good interpretation and theoretical guarantee~\cite{huang2018some}.
Deep unfolding models denote a series of models constructed by mapping iterative algorithms with unfixed numbers of steps onto deep neural networks with fixed numbers of steps~\cite{metzler2017learned,zhang2018ista,dong2018denoising,2020AMP}. There are a lot of non-linear iterative algorithms that are unfolded, such as ISTA~\cite{gregor2010learning,zhang2018ista}, AMP~\cite{metzler2017learned,2020AMP}, half-quadratic splitting (HQS)~\cite{dong2018denoising}, alternating direction methods of multipliers (ADMM)~\cite{sun2016deep,ma2019deep} and iPiano algorithm~\cite{2020iPiano}. By combining the interpretability of model-based methods and the trainable characteristics of traditional deep learning models, they make a good balance between  reconstruction performance and interpretation.

Usually, the above two kinds of deep-learning-based models are trained end-to-end using some well-known backpropagation algorithms~\cite{kingma2014adam}. However, a common shortage of most existing end-to-end-trained models is that different models have to be trained for different CS ratios. In some applications, sampling and reconstructing images at different CS ratios may be required. For examples, in image CS for wireless broadcast~\cite{Yin2016Compressive,li2013new}, users reconstruct images with different quality according to different channel conditions. And in SPI~\cite{kulkarni2016reconnet}, users can apply different CS ratios for different image quality. However, storing more than one model with the same structure would bring additional burdens to the hardware. Thus, sampling and reconstructing images at different CS ratios with only one model is needed.

\begin{figure*}
	\begin{center}
		\includegraphics{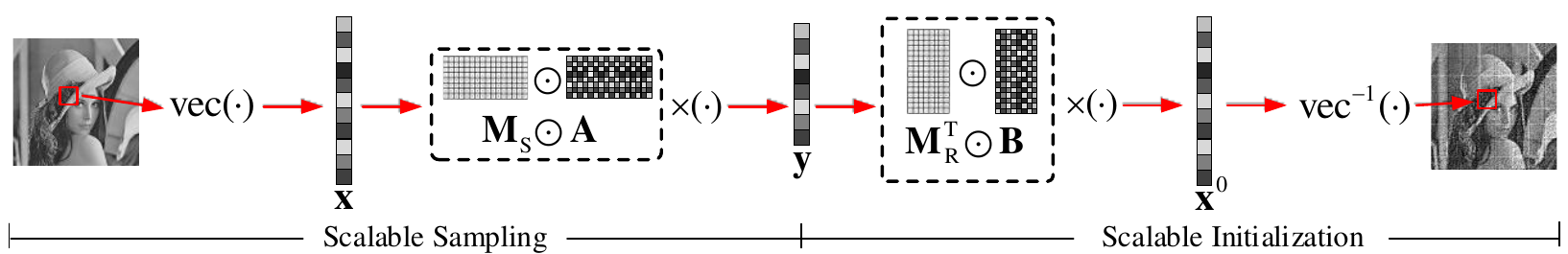}
	\end{center}
	\caption{Scalable sampling and scalable initialization of an image block.}
	\label{fig:sampling_initialization}
\end{figure*}

At present, there exist a few methods ~\cite{xu2020compressed, 2020iPiano, shi2019scalable, lohit2018rate,li2020scalable} which reconstruct images at different CS ratios using only one model, and they can be roughly cast into two categories. The first kind~\cite{xu2020compressed, 2020iPiano} trains a single model with a set of sampling matrices with different CS ratios so that the model can adapt to all sampling matrices in this set. The second kind~\cite{shi2019scalable, lohit2018rate} applies only one sampling matrix in a learning way and integrates its rows to achieve sampling and reconstruction at different CS ratios, and we call such a strategy as {\bf{s}}calable {\bf{s}}ampling and {\bf{r}}econstruction (SSR) in this paper.
And in this paper, we focus on SSR methods, because they are more practical in existing applications such as MRI~\cite{sun2016deep}, SPI~\cite{kulkarni2016reconnet} and wireless broadcasting~\cite{Yin2016Compressive}, and it has been proved that the trained sampling matrix can improve the reconstruction performance~\cite{shi2019image,2020AMP,iliadis2020deepbinarymask}. However, existing SSR methods cannot be applied universally~\cite{shi2019scalable} or a more appropriate sampling matrix is needed~\cite{lohit2018rate}. Therefore, a general and more effective SSR method is expected.

In this paper, inspired by some methods~\cite{shi2019image,2020AMP} which jointly train the sampling matrices with the models, we propose a general framework dubbed {\bf{s}}calable {\bf d}eep {\bf{c}}ompressive {\bf{s}}ensing (SDCS) to achieve sampling and reconstructing images at all CS ratios in a certain range.
In detail, a trainable initialization matrix is designed to map measurements to their original shapes.
And two binary sampling matrix masks are introduced to control the CS ratios for SSR.
Most importantly, a novel training strategy named scalable training is developed to obtain an appropriate combination of sampling matrix and model, which performs well at all CS ratios in a certain range.
We emphasize that SDCS can bring the model the ability of SSR, while maintaining the characteristics of its own structure. Furthermore, experimental results show that the model with SDCS can obtain a more effective combination of sampling matrix and model than existing SSR methods.

Our paper has three contributions:
\begin{itemize}
	\item We propose a framework named SDCS that jointly trains the sampling matrix and the model to achieve sampling and reconstruction at all CS ratios in a certain range.
	
	\item With SDCS, a deep learning model can achieve SSR without changing its original structure, while maintaining good performance.
	
	\item Technically, SDCS can be used for all end-to-end-trained deep learning models.
\end{itemize}

The framework of the remaining content is as follows: Section II describes the proposed framework SDCS. Section III introduces some related works of this paper. Section IV is the experimental results. Section V concludes this paper.

\begin{figure*}
	\begin{center}
		\includegraphics{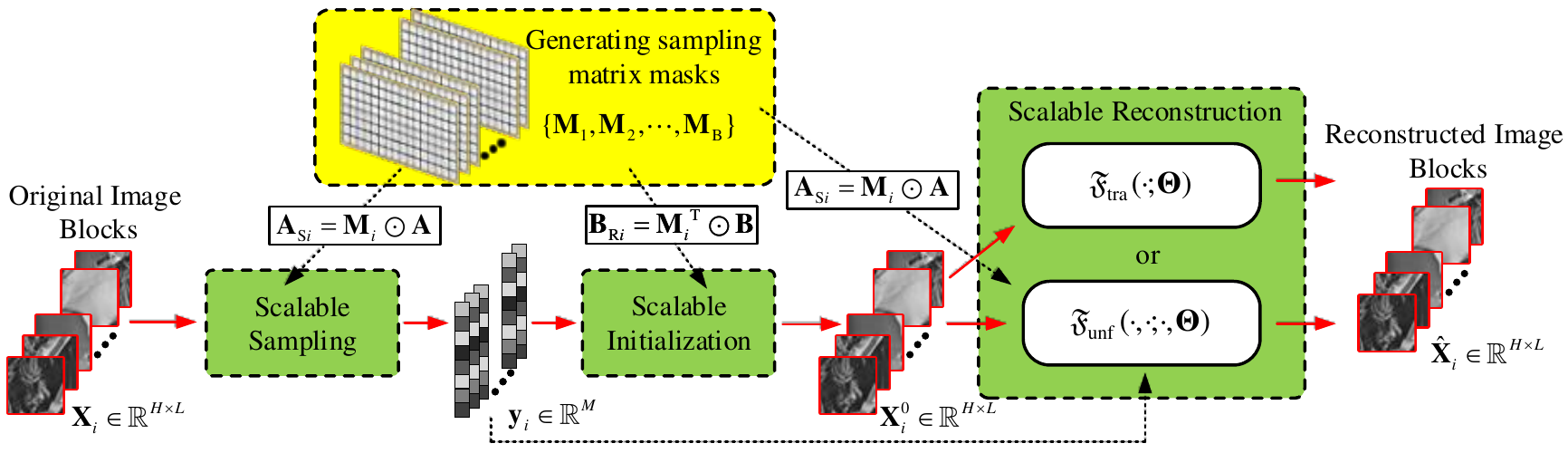}
	\end{center}
	\caption{Forward-propogation of the scalable training.}
	\label{fig:scalable_training}
	\vspace{-3mm}
\end{figure*}

\section{SDCS}
In this section, we introduce the proposed framework SDCS which is simple but  powerful. SDCS is composed of four parts: scalable sampling, scalable initialization, scalable reconstruction and scalable training.

\subsection{Scalable Sampling}
Assume that the largest CS ratio is $R_{\text{M}}$, then the sampling matrix can be expressed as $\mathbf{A}\in {\mathbb{R}^{\left\lceil {{R_{\text{M}}}N} \right\rceil \times N}}$. It can be noticed that the CS ratio is determined by the row number of $\mathbf{A}$. 
Therefore, to achieve scalable sampling, we design a sampling matrix mask $\mathbf{M}_{\text{S}}\in {\mathbb{R}^{\left\lceil {{R_{\text{M}}}N} \right\rceil \times N}}$ to control the activities of the rows of $\mathbf{A}$. $\mathbf{M}_{\text{S}}$ is a zero-one matrix which satisfies $\mathbf{M}_{\text{S}}(1:\left\lceil {{R_{\text{S}}}N} \right\rceil,:)=1$ and $\mathbf{M}_{\text{S}}(\left\lceil {{R_{\text{S}}}N} \right\rceil+1:\left\lceil {{R_{\text{M}}}N} \right\rceil,:)=0$, where $R_{\text{S}}$ denotes the CS ratio for sampling.
In such a case, we can generate a new sampling matrix as  $\mathbf{A}_{\text{S}}=\mathbf{M}_{\text{S}} \odot \mathbf{A}$, where $\odot$ denotes the element-wise product. 
Since the $\left\lceil {{R_{\text{S}}}N} \right\rceil+1$-th row to the $\left\lceil {{R_{\text{M}}}N} \right\rceil$-th row of $\mathbf{A}_{\text S}$ are all filled with 0,
we say that the first row to the $\left\lceil {{R_{\text{S}}}N} \right\rceil$-th row of $\mathbf{A}$ are activated.
In detail, if the original image block is $\bar{\mathbf{X}} \in \mathbb{R}^{H \times L}$ satisfying $N=HL$,  then the scalable sampling at the CS ratio of $R_{\text{S}}$ can be expressed as:
\begin{align}
\label{equ:scalable_sampling}
{\mathbf{y}} = \mathbf{A}_{\text{S}}\operatorname{vec}({\bar{\mathbf{X}}}),
\end{align}
where $\operatorname{vec}(\cdot)$ is an operator which transforms a matrix to a vector and $\mathbf{y} \in {\mathbb{R}^{\left\lceil {{R_{\text{M}}}N} \right\rceil}}$ is the measurement. It can be noticed that $\mathbf{y}(\left\lceil {{R_{\text{S}}}N} \right\rceil+1:\left\lceil {{R_{\text{M}}}N} \right\rceil)=0$ and $\mathbf{y}(1:\left\lceil {{R_{\text{S}}}N} \right\rceil)$ is the valid measurement for reconstruction.


\subsection{Scalable Initialization}
For deep learning methods, the initialized image is important in the following reconstruction. In SDCS, we use a linear operation to initialize the image block. 

An initialization matrix $\mathbf{B} \in {\mathbb{R}^{N \times \left\lceil {{R_{\text{M}}}N} \right\rceil}}$ is developed. Similar to (\ref{equ:scalable_sampling}),
a sampling matrix mask $\mathbf{M}_{\text R} \in {\mathbb{R}^{\left\lceil {{R_{\text{M}}}N} \right\rceil \times N}}$ is proposed to control the activities of the columns of $\mathbf{B}$, where $\mathbf{M}_{\text{R}}(1:\left\lceil {{R_{\text{R}}}N} \right\rceil,:)=1$ and $\mathbf{M}_{\text{R}}(\left\lceil {{R_{\text{R}}}N} \right\rceil+1:\left\lceil {{R_{\text{M}}}N} \right\rceil,:)=0$. $R_{\text{R}}$ denotes the CS ratio for initialization and reconstruction, which satisfies $R_{\text{R}} \leq R_{\text{S}}$.
In such a case, we can activate the first column to the $\left\lceil {{R_{\text{R}}}N} \right\rceil$-th column of $\mathbf{B}$ to generate a new initialization matrix  as $\mathbf{B}_{\text{R}}={\mathbf{M}_{\text R}}^{\operatorname{T}} \odot \mathbf{B}$. The detailed scalable initialization at the CS ratio of $R_{\text{R}}$ can be expressed as:
\begin{align}
\label{equ:scalable_Initialization}
\mathbf{X}^{0} = \operatorname{vec}^{-1}(\mathbf{B}_{\text{R}}\mathbf{y}),
\end{align}
where $\mathbf{X}^{0} \in \mathbb{R}^{H \times L}$ denotes the initialized image block and $\operatorname{vec}^{-1}(\cdot)$ is the operator which transforms a vector to matrix. 

In some cases, $R_{\text R}$ can be lower than $R_{\text S}$. For example, in wireless broadcasting~\cite{Yin2016Compressive}, images are transferred at a high CS ratio and are received at a low CS ratio due to the poor channel condition.
Fig. \ref{fig:sampling_initialization} illustrates the scalable sampling and scalable initialization of an image. 

\begin{algorithm}[t]
	\caption{Scalable training of one epoch.} 
	\label{alg:scalable_training}
	\hspace*{0.02in} {\bf Input:} 
	training set $\mathbb{T}$, batch size $B$, max CS ratio $R_{\text{M}}$, sampling matrix $\mathbf{A}$, initialization matrix $\mathbf{B}$, reconstruction model $\mathfrak{F}_{\text{tra}}(\cdot;\mathbf{\Theta})$ or $\mathfrak{F}_{\text{unf}}(\cdot;\mathbf{A},\mathbf{\Theta})$.\\
	\hspace*{0.02in} {\bf Output:} 
	trained parameters.
	\begin{algorithmic}[1]
		\State ${\mathbb{T}}' \leftarrow \emptyset $
		\Repeat
		\State Select $\mathbb{S}=\{\mathbf{X}_1, \mathbf{X}_2, \cdots, \mathbf{X}_{B}\}$ $\in \mathbb{T} \setminus \mathbb{T}'$.
		\State $\mathbb{T}' \leftarrow \mathbb{T}' \cup \mathbb{S}$.
		\State Generate $\{R_1, R_2, \cdots, R_B\}$ randomly, where $R_i \in [1, R_{\text{M}}]$.
		\State Generate $\{ \mathbf{M}_1, \mathbf{M}_2, \cdots, \mathbf{M}_B\}$, where $\mathbf{M}_i(1:\left\lceil {R_iN} \right\rceil, :)=1$ and $\mathbf{M}_i(\left\lceil {R_iN} \right\rceil+1:\left\lceil {{R_{\text{M}}}N}\right\rceil, :)=0$.
		\State Generate $\mathbb{A}_{\text S}=\{ \mathbf{A}_{\text{S}1}, \mathbf{A}_{\text{S}2}, \cdots, \mathbf{A}_{\text{S}B}\}$, $\mathbb{A}_{\text R}=\{ \mathbf{A}_{\text{R}1}, \mathbf{A}_{\text{R}2}, \cdots, \mathbf{A}_{\text{R}B}\}$ and $\mathbb{B}_{\text R}=\{ \mathbf{B}_{\text{R}1}, \mathbf{B}_{\text{R}2}, \cdots, \mathbf{B}_{\text{R}B}\}$, where $\mathbf{A}_{\text{S}i}=\mathbf{M}_i \odot \mathbf{A}$, $\mathbf{A}_{\text{R}i}=\mathbf{M}_i \odot \mathbf{A}$ and $\mathbf{B}_{\text{R}i}= \mathbf{M}^{\operatorname{T}}_i \odot \mathbf{B}$.
		\For{$i=1:B$}
		\State $\mathbf{y}_{i}=\mathbf{A}_{\text{S}i}\operatorname{vec}(\mathbf{X}_i)$
		\State $\mathbf{X}_{i}^{0}=\operatorname{vec}^{-1}(\mathbf{B}_{\text{R}i}\mathbf{y}_i)$
		\State $\hat{\mathbf{X}}_{i} =  \mathfrak{F}_{\text{tra}}(\mathbf{X}^{0}_{i};\mathbf{\Theta}) $ or $\hat{\mathbf{X}}_{i}=\mathfrak{F}_{\text {unf}}(\mathbf{X}^{0}_{i},\mathbf{y}_{i};\mathbf{A}_{\text{R}i},\mathbf{\Theta})$
		\EndFor
		\State Compute loss $L$ using  $\{\hat{\mathbf{X}}_{1},\hat{\mathbf{X}}_{2},\cdots,\hat{\mathbf{X}}_{B}\}$ and $\mathbb{S}$.
		\State Updating $\mathbf{A}$, $\mathbf{B}$ and $\mathbf{\Theta}$.
		\Until{$\mathbb{T} \setminus \mathbb{T}'=\emptyset$}
		\State \Return $\mathbf{A}$, ${\mathbf{B}}$, $\mathbf{\Theta}$.
	\end{algorithmic}
\end{algorithm}

\subsection{Scalable Reconstruction}
In this subsection, we describe the scalable reconstruction of two different kinds of deep learning models: traditional deep learning models and deep unfolding models. 

The generalized reconstruction process of traditional deep learning models can be expressed as:
\begin{align}
\label{equ:tradition_reconstruction}
{\hat{\mathbf{X}}} =  \mathfrak{F}_{\text{tra}}(\mathbf{X}^{0};\mathbf{\Theta}),
\end{align}
where ${\hat{\mathbf{X}}} \in \mathbb{R}^{H \times L}$ is the reconstructed image block and $\mathbf{\Theta}$ contains trainable parameters of the model. In SDCS, $\mathbf{\Theta}$ is trained with $\mathbf{A}$ and $\mathbf{B}$ to make sure that $\mathfrak{F}_{\text{tra}}(\cdot;\mathbf{\Theta})$ can perform well at all CS ratios.

The reconstruction model of a deep unfolding model is usually composed of $K$ reconstruction modules with the same structure. 
In each module, the sampling matrix also participates in the image reconstruction. 
In detail, the generalized reconstruction process of a deep unfolding model can be expressed as:
\begin{align}
\label{equ:unfolding_reconstruction}
\hat{\mathbf{X}} &= \mathfrak{F}_{\text {unf}}(\mathbf{X^{0}},\mathbf{y};\mathbf{A},\mathbf{\Theta})=
\mathfrak{F}_{\text{unf}}^K(\mathbf{X}^{K-1},\mathbf{y};\mathbf{A},\mathbf{\Theta}^{K}),\\
\mathbf{X}^{k}&=
\mathfrak{F}_{\text{unf}}^k(\mathbf{X}^{k-1},\mathbf{y};\mathbf{A},\mathbf{\Theta}^{k}),
\end{align}
where $\mathfrak{F}_{\text {unf}}(\cdot,\cdot;\mathbf{A},\mathbf{\Theta})$ is the entire deep unfolding model, of which $\mathbf{\Theta}$ contains its trainable parameters. $\mathfrak{F}_{\text {unf}}^k(\cdot,\cdot;\mathbf{A},\mathbf{\Theta}^{k})$ is the $k$-th reconstruction module and $\mathbf{\Theta}^{k}$ contains its trainable parameters.
the inputs of $\mathfrak{F}_{\text {unf}}(\cdot,\cdot;\mathbf{A},\mathbf{\Theta})$ and $\mathfrak{F}_{\text {unf}}^k(\cdot,\cdot;\mathbf{A},\mathbf{\Theta}^{k})$ usually contain the image block $\mathbf{X}^0$ and the measurement $\mathbf{y}$.
Since $\mathbf{A}$ plays an important role in each reconstruction module, the scalable reconstruction of the deep unfolding model is achieved by applying activated sampling matrix. In detail, the scalable reconstruction of the $k$-th reconstruction model can be expressed as:
\begin{align}
\label{equ:scalable_unfolding_reconstruction}
\mathbf{X}^{k}&=
\mathfrak{F}_{\text{unf}}^k(\mathbf{X}^{k-1},\mathbf{y};{\mathbf{M}_{\text R}} \odot \mathbf{A},\mathbf{\Theta}^{k}).
\end{align}
Similar the traditional deep learning models, $\mathbf{\Theta}=\{\mathbf{\Theta}^{1}, \mathbf{\Theta}^{2}, \cdots, \mathbf{\Theta}^{K}\}$ is trained with $\mathbf{A}$ and $\mathbf{B}$.
Since the sampling matrix $\mathbf{A}$ usually appears in the image sampling and reconstruction of deep unfolding models, deep unfolding models have a great potential to achieve SSR.

\subsection{Scalable Training}
As shown in (\ref{equ:scalable_sampling}), (\ref{equ:scalable_Initialization}) and (\ref{equ:scalable_unfolding_reconstruction}), $\mathbf{A}$ and $\mathbf{B}$ are important in effective SSR. How to obtain an appropriate combination of $\mathbf{A}$, $\mathbf{B}$ and the reconstruction model is the main issue.
To this end, we develop a novel training strategy dubbed scalable training to train $\mathbf{A}$, $\mathbf{B}$ with parameters of the reconstruction model jointly.

In scalable training, it is assumed that all parameters are trained using stochastic-gradient-descent-related algorithms like Adam~\cite{kingma2014adam}. If the batch size for training is $B$, the training process of $\mathbf{A}$, $\mathbf{B}$ and $\mathbf{\Theta}$ of one epoch can be expressed as Algorithm \ref{alg:scalable_training}. And Fig. \ref{fig:scalable_training} illustrates the forward-propagation of the scalable training. The gradients of $\mathbf{A}$ and $\mathbf{B}$ can be computed as follows:
\begin{align}
\label{equ:gradient_A}
{\nabla _{\mathbf A}}L = \frac{1}{B}\sum\limits_{i = 1}^B {{{\mathbf{M}}_{i}} \odot } {\nabla _{{{\mathbf{M}}_{i}} \odot \mathbf{A}}}L,\\
\label{equ:gradient_B}
{\nabla _{\mathbf B}}L = \frac{1}{B}\sum\limits_{i = 1}^B {{{\mathbf{M}}_{i}}^{\operatorname{T}} \odot } {\nabla _{{{\mathbf{M}}_{i}}^{\operatorname{T}} \odot \mathbf{B}}}L.
\end{align}
It can be noticed that the closer to the top of $\mathbf{A}$ or the left of $\mathbf{B}$, the more gradient information for updating is obtained, which makes using $\mathbf{M}_{\text S}$ and $\mathbf{M}_{\text R}$ for effective SSR possible.

Furthermore, to validate the trained model, a CS {\bf r}atio {\bf v}alidation {\bf g}roup (RVG) is applied. Each RVG contains $G$ validation CS ratios as $\{{R_1},{R_2}, \cdots ,{R_G}\}$. At the end of each epoch, for each ratio $R_i$, the average PSNR on the validation set can be obtained. And the model with the best average PSNR on RVG is regarded as the model for test. 

We emphasize that SDCS has no restriction on the structure of deep learning models, which means it can be combined with any end-to-end-trained model for SSR. However, the final performance is determined by the structure of the reconstruction model.

\section{Related Works}
In this section, we first introduce some deep-learning-based methods for image CS, then some SSR methods are compared with SDCS.

\subsection{Deep Learning Models for Image CS}
For traditional deep learning models, Mousavi et al.~\cite{mousavi2015deep} first designed a fully-connected-layer-based stacked denoising autoencoder (SDA) for visual image CS. Lohit et al.~\cite{kulkarni2016reconnet} first proposed a six-layers CNN-based model named ReconNet to reconstruct image blocks from measurements. Shi et al.~\cite{shi2019image} proposed a deeper CNN model named CSNet which has trainable deblocking operations and integrated residual connection~\cite{he2016deep} for better performance. Furthermore, there are some other models~\cite{du2019fully,yao2019dr2,pmlr-v70-bora17a,sun2020learning} for image CS, and all these models have one thing in common that the models for reconstruction are trained end-to-end.

Deep unfolding models are first developed for the sparse coding problem~\cite{gregor2010learning,chen2018theoretical,borgerding2017amp}. And inspired by these models, Zhang et al.~\cite{zhang2018ista} developed a deep unfolding model named ISTA-Net for image CS problem by unfolding ISTA and learning sparse transformation functions. Metzler et al.~\cite{metzler2017learned} and Zhang et al.~\cite{2020AMP} established deep unfolding models named LDAMP and AMP-Net respectively based on AMP algorithm, where LDAMP samples and reconstructs the entire image, and AMP-Net measures and recovers an image block-by-block with general trainable deblocking modules. Dong et al.~\cite{dong2018denoising} designed an model named DPDNN inspired by HQS algorithm for image inverse problem which can be applied for image CS. These deep unfolding models apply the sampling matrix for reconstruction and they can also be trained end-to-end.

Some of the above methods discuss the sampling matrix training strategies, including in traditional deep learning models~\cite{mousavi2015deep,shi2019image} or in deep unfolding model~\cite{2020AMP}. Although the trained sampling  matrices can improve the reconstruction performance, they are designed for the single CS ratios and the performance would decrease seriously when the CS ratio changes for SSR. However, using SDCS, the model and the trained sampling matrix can perform well in all CS ratios in a certain range.

\subsection{SSR Methods}
\label{related_works_SSR_methods}
As far as we know, there exist two SSR methods~\cite{shi2019scalable,lohit2018rate} for the visual image CS. Shi et al.~\cite{shi2019scalable} proposed a model dubbed SCSNet and Lohit et al.~\cite{lohit2018rate} designed a general framework named Rate-Adaptive CS (RACS). We compare them with SDCS in the following two paragraphs.

SCSNet trains the sampling matrix with the reconstruction model which is composed of seven independent sub-models with the same structure. Each sub-model adapts to a sub-range of CS ratios to make sure that the whole model can achieve SSR at CS ratios from 1\% to 50\%. And a greedy algorithm is applied to rearrange the rows of the sampling matrix for better reconstruction. However, SCSNet has two weaknesses: 1) The number of parameters is very large due to the existence of multiple sub-models. 2) Based
on SCSNet, the existing deep learning models have to change their structure to achieve scalable reconstruction which would bring more burden to the hardware. However, SDCS needs only one model to achieve SSR and it can be applied for all end-to-end-trained models without changing their structures.

\begin{table*}
	\begin{center}
		\footnotesize
		\caption{The results of twelve models tested on Set11 at different CS ratios, where the best is marked in bold.}
		\begin{tabular}{|l|c|c|c|c|c|c|c|}
			\hline
			\multirow{2}*{{\bf Method}} & 50$\%$ & 40$\%$ &30$\%$ &25$\%$ &10$\%$ &4$\%$ &1$\%$\\
			\cline{2-8}
			& \multicolumn{7}{c|}{PSNR (dB)/SSIM}\\
			\hline\hline
			
			SDA~\cite{mousavi2015deep} & 26.43/0.8007  &25.14/0.7371 & 24.77/0.7191 & 24.77/0.7234 & 23.66/0.6794 & 21.05/0.5720 & 17.69/0.4376 \\ 
			
			SDA-SDCS & {\bf30.80}/{\bf0.9038}  &{\bf30.63}/{\bf0.9009} & {\bf29.43}/{\bf0.8793} & {\bf28.76}/{\bf0.8636} & {\bf25.58}/{\bf0.7660} & {\bf22.77}/{\bf0.6458} & {\bf19.87}/{\bf0.4829}\\
			\hline
			
			ReconNet~\cite{kulkarni2016reconnet} & 32.12/0.9137 &30.59/0.8928 & 28.72/0.8517 & 28.04/0.8303 & 24.07/0.6958 & 21.00/0.5817 & 17.54/0.4426\\
			
			ReconNet-SDCS & {\bf34.29}/{\bf0.9532}  &{\bf33.81}/{\bf0.9242} & {\bf32.42}/{\bf0.9313} & {\bf31.42}/{\bf0.9173} & {\bf26.90}/{\bf0.8225} & {\bf23.57}/{\bf0.6931} & {\bf20.02}/{\bf0.5071}\\
			\hline
			
			${\text{CSNe}}{{\text{t}^+}}$~\cite{shi2019image} & {\bf38.19}/{\bf0.9739} &{\bf36.15}/{\bf0.9625} & {\bf33.90}/{\bf0.9449} & {\bf32.76}/{\bf0.9322} & 27.76/{\bf0.8513} & {\bf24.24}/{\bf0.7412} & 20.09/0.5334\\
			
			${\text{CSNe}}{{\text{t}^+}}$-SDCS&36.65/0.9645&35.48/0.9568&33.58/0.9414&32.44/0.9295&{\bf27.85}/0.8493&23.92/0.7303&{\bf20.32}/{\bf0.5394}\\
			\hline
			
			${\text{ISTA-Ne}}{{\text{t}}^+}$~\cite{zhang2018ista} & {\bf38.08}/0.9680 &{\bf35.93}/0.9537 & {\bf33.66}/0.9330 & {\bf32.27}/0.9167 & 25.93/0.7840  & 21.14/0.5947 & 17.48/0.4403\\
			
			${\text{ISTA-Ne}}{{\text{t}}^+}$-SDCS & 36.51/{\bf0.9693}  &{34.92}/{\bf0.9587} & 32.85/{\bf0.9400} & {31.65}/{\bf0.9256} & {\bf26.99}/{\bf0.8334} & {\bf23.57}/{\bf0.7073} & {\bf20.13}/{\bf0.5146}\\ 
			\hline
			
			DPDNN~\cite{dong2018denoising}&35.85/0.9532&34.30/0.9411&32.06/0.9145&30.63/0.8924&24.53/0.7392&21.11/0.6029&17.59/0.4459\\
			
			DPDNN-SDCS & {\bf39.50}/{\bf0.9775} &{\bf37.61}/{\bf0.9686} & {\bf35.38}/{\bf0.9543} & {\bf34.12}/{\bf0.9434} & {\bf29.07}/{\bf0.8708} & {\bf25.08}/{\bf0.7622} & {\bf20.55}/{\bf0.5423}\\
			\hline
			
			AMP-Net~\cite{2020AMP} & {\bf40.27}/{\bf0.9804} &{\bf38.23}/{\bf0.9713} & {\bf35.90}/{0.9574} & {34.59}/{\bf0.9477} & {29.45}/{0.8787} & {25.16}/{0.7692} & {\bf20.57}/{\bf0.5639}\\
			
			AMP-Net-SDCS&39.67/0.9781&37.96/0.9703&35.89/{\bf0.9576}&{\bf34.67}/0.{\bf9477}&{\bf29.59}/{\bf0.8792}&{\bf25.43}/{\bf0.7750}&{20.47}/{0.5629}\\
			\hline
		\end{tabular}
		\label{tab:Set11}
	\end{center}
\end{table*}

\begin{table*}
	\begin{center}
		\footnotesize
		\caption{The results of twelve models tested on the test set of BSDS500 at different CS ratios, where the best is marked in bold.}
		\begin{tabular}{|l|c|c|c|c|c|c|c|}
			\hline
			\multirow{2}*{{\bf Method}} & 50$\%$ & 40$\%$ &30$\%$ &25$\%$ &10$\%$ &4$\%$ &1$\%$\\
			\cline{2-8}
			& \multicolumn{7}{c|}{PSNR (dB)/SSIM}\\
			\hline\hline
			
			SDA~\cite{mousavi2015deep} & 26.16/0.8048  &24.97/0.7392 & 24.58/0.7127 & 24.58/0.7107 & 23.77/0.6489 & 21.75/0.5534 & 19.05/0.4522 \\ 
			
			SDA-SDCS & {\bf30.17}/{\bf0.9026}  &{\bf29.90}/{\bf0.8973} & {\bf28.77}/{\bf0.8704} & {\bf28.13}/{\bf0.8510} & {\bf25.43}/{\bf0.7338} & {\bf23.38}/{\bf0.6145} & {\bf21.08}/{\bf0.4865}\\
			\hline
			
			ReconNet~\cite{kulkarni2016reconnet} & 30.85/0.8949 &29.47/0.8647 & 27.95/0.8190 & 27.20/0.7914 & 23.98/0.6472 & 21.69/0.5557 & 18.96/0.4531\\
			
			ReconNet-SDCS & {\bf33.27}/{\bf0.9448}  &{\bf32.52}/{\bf0.9355} & {\bf31.04}/{\bf0.9107} & {\bf30.13}/{\bf0.8921} & {\bf26.46}/{\bf0.7753} & {\bf23.99}/{\bf0.6502} & {\bf21.20}/{\bf0.5063}\\
			\hline
			
			${\text{CSNe}}{{\text{t}^+}}$~\cite{shi2019image} & {\bf35.89}/{\bf0.9677} &{\bf33.96}/{\bf0.9513} & {\bf31.94}/{\bf0.9251} & {\bf30.91}/{\bf0.9067} & {\bf27.01}/{\bf0.7949} & {\bf24.41}/{\bf0.6747} & {21.42}/{0.5261}\\
			
			${\text{CSNe}}{{\text{t}^+}}$-SDCS&34.91/0.9588&33.59/0.9462&31.80/0.9221&30.82/0.9043&26.97/0.7906&24.21/0.6692&{\bf21.48}/{\bf0.5288}\\
			\hline
			
			${\text{ISTA-Ne}}{{\text{t}}^+}$~\cite{zhang2018ista} & {\bf34.92}/0.9510 &32.87/0.9264 & 30.77/0.8901 & 29.64/0.8638 & 25.11/0.7124  & 21.82/0.5661 & 18.92/0.4529\\
			
			${\text{ISTA-Ne}}{{\text{t}}^+}$-SDCS & 34.85/{\bf0.9622}  &{\bf33.26}/{\bf0.9465} & {\bf31.38}/{\bf0.9199} & {\bf30.36}/{\bf0.9003} & {\bf26.56}/{\bf0.7811} & {\bf24.00}/{\bf0.6555} & {\bf21.24}/{\bf0.5096}\\ 
			\hline
			
			DPDNN~\cite{dong2018denoising}&33.56/0.9373&32.05/0.9164&29.98/0.8759&28.87/0.8491&24.37/0.6863&21.80/0.5716&18.97/0.4544\\
			
			DPDNN-SDCS & {\bf36.84}/{\bf0.9708} &{\bf34.91}/{\bf0.9560} & {\bf32.85}/{\bf0.9323} & {\bf31.74}/{\bf0.9150} & {\bf27.58}/{\bf0.8069} & {\bf24.78}/{\bf0.6858} & {\bf21.72}/{\bf0.5319}\\
			\hline
			
			AMP-Net~\cite{2020AMP} & {\bf37.48}/{\bf0.9744} &{\bf35.34}/{\bf0.9594} & {\bf33.17}/{\bf0.9358} & {32.01}/{\bf0.9188} & {27.82}/{0.8133} & {24.95}/{0.6949} & {\bf21.90}/{\bf0.5501}\\
			
			AMP-Net-SDCS&37.04/0.9720&35.18/0.9580&33.14/0.9354&{\bf32.04}/0.9187&{\bf27.84}/{\bf0.8136}&{\bf25.03}/{\bf0.6967}&{21.87}/{0.5493}\\
			\hline
		\end{tabular}
		\label{tab:BSD500}
	\end{center}
\end{table*}

\begin{table*}
	\setlength{\belowcaptionskip}{0pt}
	\setlength{\abovecaptionskip}{2pt}
	\begin{center}
		\footnotesize
		\caption{Parameter number of the reconstruction model of seven models.}
		\begin{tabular}{|c|c|c|c|c|c|c|c|}
			\hline
			{{ Parameter}} & SDA-SDCS& ReconNet-SDCS&${\text{CSNe}}{{\text{t}^+}}$-SDCS&${\text{ISTA-Ne}}{{\text{t}}^+}$-SDCS&DPDNN-SDCS&AMP-Net-SDCS&SCSNet\\
			\cline{2-8}
			{{ Number}}&6534&22914&370560&336978&1363712&229254&1110823\\
			\hline
		\end{tabular}
		\label{tab:PN}
	\end{center}
	\vspace{-5mm}
\end{table*}

RACS is a general framework like SDCS and it has three training stages. In the stage 1, the model is trained with the sampling matrix at a single CS ratio of $R_{\text M}$. And all parameters of the model are frozen after the stage 1. In the stage 2, The first $R_{\text K}N$ rows of the sampling matrix are optimized, where $R_{\text K}<R_{\text M}$. In the stage 3, the following rows of the sampling matrix are trained one-by-one. It can be noticed that RACS has an obvious weakness: the model is learned for a specific sampling matrix with CS ratio $R_{\text M}$ in the stage 1, which means the performance of the model at lower CS ratios can be further improved.
Different from RACS, with SDCS, the learned model adapt to a sampling matrix which can change its CS ratios from 1\% to $R_{\text M}$ using a sampling matrix mask. Our strategy brings the model the potential that performs better for SSR.

\section{Experimental Results}
\subsection{Experimental settings}
In this paper, the model combined with SDCS is named as \emph{model}-SDCS.
To evaluate the performance of SDCS, six models are combined with SDCS, namely SDA~\cite{mousavi2015deep}, ReconNet~\cite{kulkarni2016reconnet}, ${\text{CSNe}}{{\text{t}^+}}$~\cite{shi2019image}, ${\text{ISTA-Ne}}{{\text{t}}^+}$~\cite{zhang2018ista}, DPDNN~\cite{dong2018denoising} and AMP-Net~\cite{2020AMP}, which sample and reconstruct images block-by-block with the block size of $33 \times 33$ that makes $N=1089$. SDA, ReconNet, ${\text{CSNe}}{{\text{t}^+}}$ are traditional deep learning models.
${\text{ISTA-Ne}}{{\text{t}}^+}$, DPDNN and AMP-Net are deep unfolding models with 9, 6 and 6 reconstruction modules respectively. In this paper, the activation fucntions of SDA are changed to the Rectified Linear Unit (ReLU)~\cite{kulkarni2016reconnet} for better performance.
It is worth noting that in ${\text{CSNe}}{{\text{t}^+}}$ and AMP-Net, trainable deblocking operations are applied and the sampling matrices are trained for a single CS ratio. Furthermore, since SCSNet~\cite{shi2019scalable} and RACS~\cite{lohit2018rate} can achieve SSR like $model$-SDCS, they are compared with SDCS to show the effectiveness of our framework.



All of our experiments are performed on two datasets: BSDS500~\cite{arbelaez2010contour} and Set11~\cite{kulkarni2016reconnet}. BSDS500 contains 500 colorful visual images and is composed of a training set (200 images), a validation set (100 images) and a test set (200 images). Set11~\cite{kulkarni2016reconnet} contains 11 grey-scale images. In this paper, BSDS500 is used for training, validation and testing. And Set11 is used for testing. We generate two training sets for models with and without trainable deblocking operations. 
(a) Training set 1 contains 89600 sub-images sized of $99\times99$ which are randomly extracted from the luminance components of images in the training set of BSDS500 ~\cite{shi2019image}.
(b) Training set 2 contains 195200 sub-images sized of $33\times33$ which are randomly extracted from the luminance components of images in the training set of BSDS500 ~\cite{ zhang2018ista}. 
In this paper ${\text{CSNe}}{{\text{t}^+}}$, AMP-Net, ${\text{CSNe}}{{\text{t}^+}}$-SDCS, AMP-Net-SDCS and SCSNet are trained on training set 1 due to the existence of trainable deblocking operations.
SDA, ReconNet, ${\text{ISTA-Ne}}{{\text{t}}^+}$, DPDNN, SDA-SDCS, ReconNet-SDCS, ${\text{ISTA-Ne}}{{\text{t}}^+}$-SDCS and DPDNN-SDCS are trained on training set 2. And they are trained on the conditions in their original papers.
Moreover, we use the validation set of BSDS500 for model choosing and the test set of BSDS500 for testing. 
In this paper, all sampling matrices are initialized randomly in Gaussian distribution.  $R_{\text{M}}$ is 50\% and RVG is $\{1\%, 4\%, 10\%, 25\%, 30\%, 40\%, 50\%\}$. All experiments are performed on a computer with an AMD Ryzen7 2700X CPU and an RTX2080Ti GPU.

\subsection{Comparison with original deep learning methods}
In this subsection, we compare SDA, ReconNet, ${\text{CSNe}}{{\text{t}^+}}$, ${\text{ISTA}}{{\text{-Net}^+}}$, DPDNN and AMP-Net with SDA-SDCS, ReconNet-SDCS, ${\text{CSNe}}{{\text{t}^+}}$-SDCS, ${\text{ISTA-Ne}}{{\text{t}}^+}$-SDCS, DPDNN-SDCS and AMP-Net-SDCS.
Table \ref{tab:Set11} and Table \ref{tab:BSD500} show the average PSNR and SSIM of 12 models tested on Set11 and the testing set of BSDS500 at different CS ratios respectively.
We emphasize that there are seven different models for seven different test CS ratios for the method without SDCS, and a single model is tested at different CS ratios for $model$-SDCS.

\begin{figure*}[htbp]
	\centering
	\subfigure{
		\label{fig:scalable_PSNR}
		\begin{minipage}[t]{0.5\linewidth}
			\centering
			\includegraphics[scale=1]{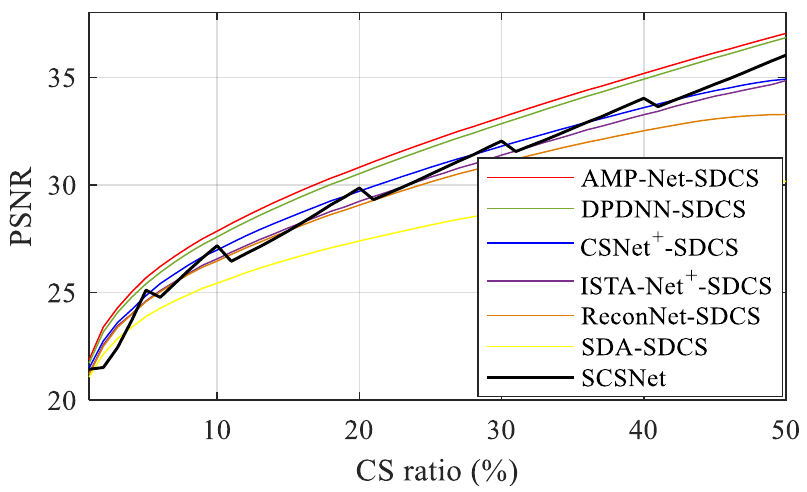}
		\end{minipage}%
	}%
	\subfigure{
		\label{fig:scalable_SSIM}
		\begin{minipage}[t]{0.5\linewidth}
			\centering
			\includegraphics[scale=1]{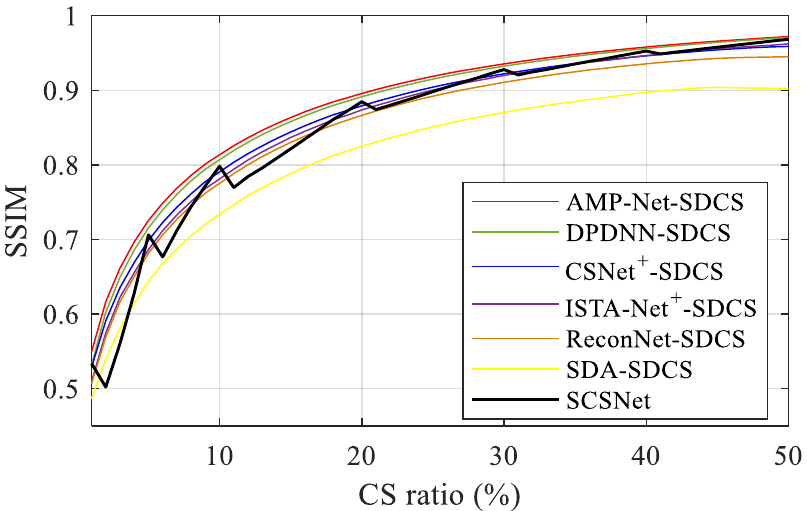}
		\end{minipage}%
	}%
	\centering
	\caption{Comparison between SCSNet and six models with SDCS at different CS ratios.}
	\label{fig:scalable}
\end{figure*}

\begin{figure*}[htbp]
	\centering
	\subfigure{
		\label{fig:RACS_psnr}
		\begin{minipage}[t]{0.5\linewidth}
			\centering
			\includegraphics[scale=1]{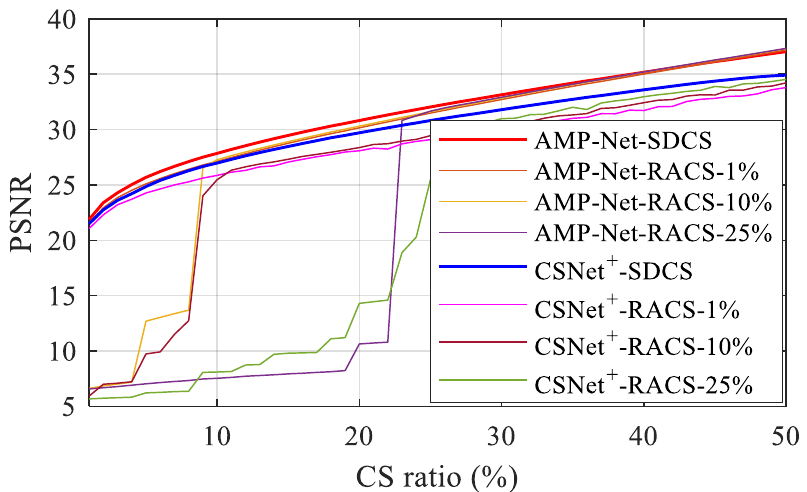}
		\end{minipage}%
	}%
	\subfigure{
		\label{fig:RACS_ssim}
		\begin{minipage}[t]{0.5\linewidth}
			\centering
			\includegraphics[scale=1]{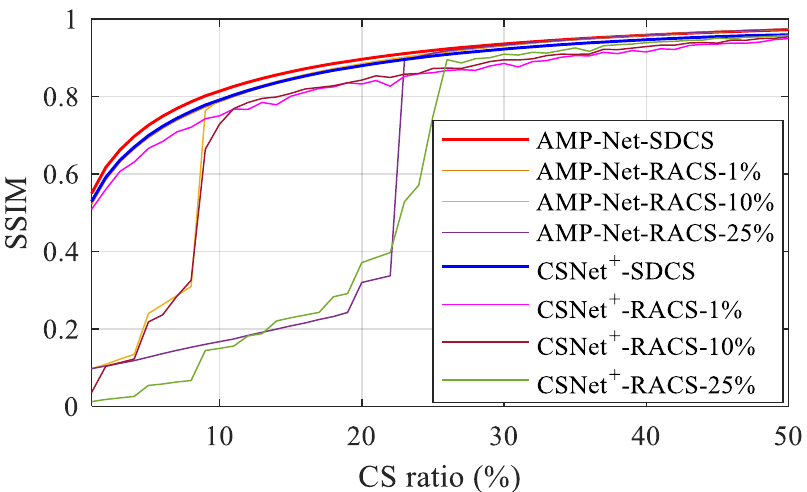}
		\end{minipage}%
	}%
	\centering
	\caption{Comparison between SDCS and RACS at different CS ratios.}
	\label{fig:RACS}
\end{figure*}




From Table \ref{tab:Set11} and Table \ref{tab:BSD500}, it can be found that 
compared with models without trained sampling matrices, although $model$-SDCS has only one model for reconstruction, it obtains better performance in terms of PSNR and SSIM at most test CS ratios.
And compared with models that apply trained sampling matrices, the average PSNR and SSIM of \emph{model}-SDCS are not much different from them at most CS ratios and even higher than them at some CS ratios. Therefore, we conclude that $model$-SDCS can effectively achieve SSR without changing the structure of the model.

Furthermore, by further analyzing ${\text{CSNe}}{{\text{t}^+}}$, ${\text{CSNe}}{{\text{t}^+}}$-SDCS, AMP-Net and AMP-Net-SDCS, it can be noticed that even the sampling matrices of ${\text{CSNe}}{{\text{t}^+}}$ and AMP-Net are trained, ${\text{CSNe}}{{\text{t}^+}}$-SDCS and AMP-Net-SDCS can still obtain competitive performance at all test CS ratios with a single model. Such a result implies the great potential of deep leaning techniques and the sampling matrix training strategy.

\subsection{Comparison with SSR methods}
In this subsection, we compare SDCS with two SSR methods: SCSNet~\cite{shi2019scalable} and RACS~\cite{lohit2018rate}.

First, SCSNet is compared with SDA-SDCS, ReconNet-SDCS, ${\text{CSNe}}{{\text{t}^+}}$-SDCS, ${\text{ISTA-Ne}}{{\text{t}}^+}$-SDCS, DPDNN-SDCS and AMP-Net-SDCS. Table \ref{tab:PN} shows the parameter number of the seven models. Fig. \ref{fig:scalable} plots average PSNR and SSIM of the seven models tested on the test set of BSDS500 at CS ratios from 1\% to 50\%.
It can be noticed that except DPDNN-SDCS, other models have fewer parameters than SCSNet and achieve SSR. And DPDNN-SDCS and AMP-Net-SDCS even outperform SCSNet, which shows the great potential of SDCS.
Furthermore, deep unfolding models have better SSR performance than traditional deep learning models. For examples, AMP-Net-SDCS and DPDNN-SDCS outperform SDA-SDCS, ReconNet-SDCS and ${\text{CSNe}}{{\text{t}^+}}$-SDCS, and ${\text{ISTA-Ne}}{{\text{t}}^+}$-SDCS outperform SDA-SDCS and ReconNet-SDCS. we conclude that deep unfolding models are more suitable for SSR to a certain degree due to the important role of the sampling matrix in the image reconstruction process.

\begin{figure*}
	\begin{center}
		\includegraphics{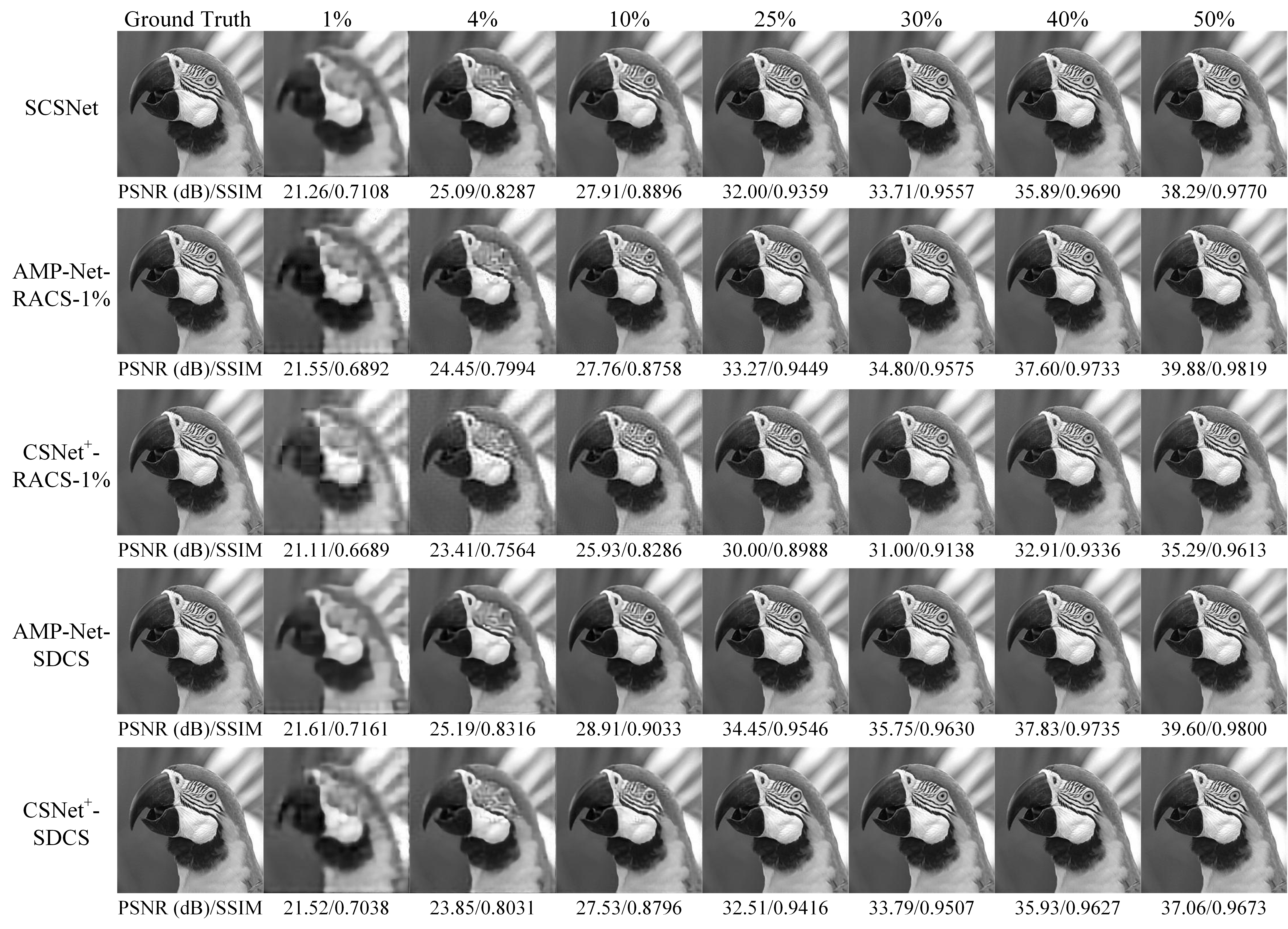}
	\end{center}
	\caption{The \textit{Parrots} images in Set11 Reconstructed by different SSR methods at different CS ratios.}
	\label{fig:reconstructed_images}
	\vspace{-3mm}
\end{figure*}

\begin{figure*}[htbp]
	\centering
	\subfigure{
		\label{fig:greedy_psnr}
		\begin{minipage}[t]{0.5\linewidth}
			\centering
			\includegraphics[scale=1]{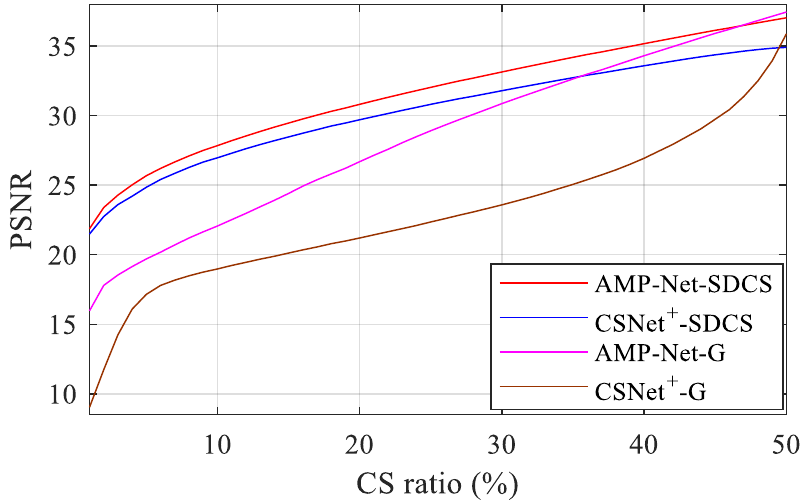}
		\end{minipage}%
	}%
	\subfigure{
		\label{fig:greedy_ssim}
		\begin{minipage}[t]{0.5\linewidth}
			\centering
			\includegraphics[scale=1]{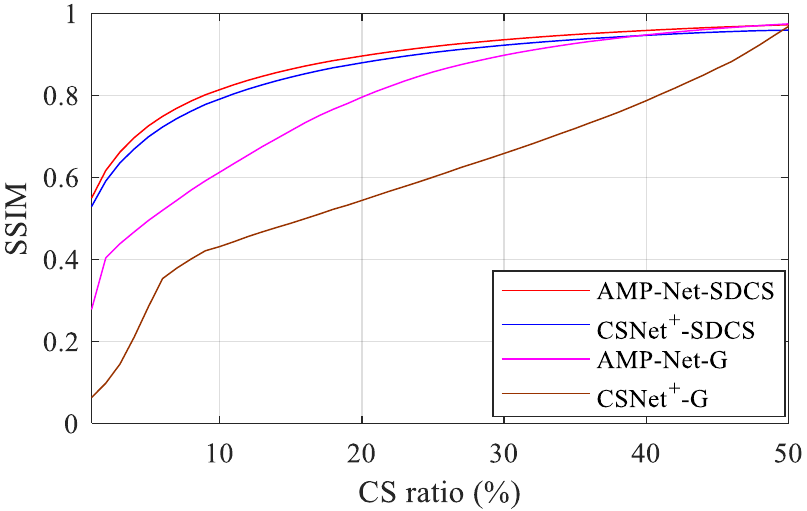}
		\end{minipage}%
	}%
	\centering
	\caption{Comparison between two models with and without SDCS at different CS ratios. $model$-G is the model combined with the greegy algorithm in \cite{shi2019scalable}.}
	\label{fig:greedy}
\end{figure*}

Second, SDCS is compared with RACS. Since with SDCS, AMP-Net outperforms other deep unfolding models and ${\text{CSNe}}{{\text{t}^+}}$ outperforms other traditional deep learning models, we use AMP-Net and ${\text{CSNe}}{{\text{t}^+}}$ as examples to compare SDCS and RACS. In this subsection, the values of $R_{\text K}$ of RACS mentioned in \ref{related_works_SSR_methods} are 1\%, 10\% and 25\%. Fig. \ref{fig:RACS} plots average PSNR and SSIM of AMP-Net-SDCS, ${\text{CSNe}}{{\text{t}^+}}$-SDCS, AMP-Net-RACS-$R_{\text K}$ and ${\text{CSNe}}{{\text{t}^+}}$-RACS-$R_{\text K}$ on the test set of BSDS500 at CS ratios from 1\% to 50\%, where $model$-RACS-$R_{\text K}$ denotes the model combined with RACS with the hyperparameter $R_{\text K}$. It can be noticed that when the CS ratio is lower than $R_{\text K}$, $model$-RACS-$R_{\text K}$ has bad performance. For AMP-Net, AMP-Net-SDCS outperforms all compared AMP-Net-RACS-$R_{\text K}$s when the CS ratio is lower than 30\%. And for ${\text{CSNe}}{{\text{t}^+}}$, ${\text{CSNe}}{{\text{t}^+}}$-SDCS has better performance than all compared ${\text{CSNe}}{{\text{t}^+}}$-RACS-$R_{\text K}$s at all CS ratios. Such a result implies that SDCS can generate a more appropriate combination of sampling matrix and model than RACS.

Fig. \ref{fig:reconstructed_images} shows the \textit{Parrots} images in Set11 reconstructed by different SSR models at different CS ratios. Fig. \ref{fig:reconstructed_images} is quite revealing in several ways. 1) AMP-Net-SDCS generates better results than SCSNet while maintaining fewer parameters which shows the great potential of SDCS. 2) $Model$-RACS-$R_{\text K}$ can not inherit the characteristics of original models well. For example, AMP-Net and CSNe${\text t}^{+}$ both have trainable deblocking operations, but AMP-Net-RACS-1\% and CSNe${\text t}^{+}$-RACS-1\% generates images with obvious blocking artifacts at CS ratios of 1\%, 4\% and 10\%. However, AMP-Net-SDCS and CSNe${\text t}^{+}$-SDCS generate smooth images without blocking artifacts. Therefore, we conclude that models with SDCS can get good SSR performance. In particular, they can inherit the characteristics of original models.

To further prove the effectiveness of SDCS, we compare AMP-Net-SDCS and ${\text{CSNe}}{{\text{t}^+}}$-SDCS with AMP-Net and ${\text{CSNe}}{{\text{t}^+}}$ which train their sampling matrices for the single CS ratio.
In this subsection, the sampling matrices of AMP-Net and ${\text{CSNe}}{{\text{t}^+}}$ are trained for the CS ratio of 50\%. And their rows are rearranged using the greedy algorithm in SCSNet~\cite{shi2019scalable} for better SSR.
Fig. \ref{fig:greedy} plots the average PSNR and SSIM of four models tested on the test set of BSDS500 at CS ratios from 1\% to 50\%. It can be noticed that at the CS ratio of 50\%, the specially trained models can obtain better results than the models with SDCS, but such models have a bad performance at other CS ratios. However, models combined with SDCS perform well at all CS ratios. 
Therefore, we conclude that compared with methods that jointly train the sampling matrix and the model for a single CS ratio, $model$-SDCS can get better SSR performance.

\section{Conclusion}
In this paper, for the visual image CS problem, we propose a general framework named SDCS to achieve sampling and reconstructing images using one sampling matrix and one deep learning model at different CS ratios. Theoretically, SDCS can be combined with all end-to-end-trained deep learning models which do not need to change their structures. Experimental results show that models with SDCS inherit the characteristics of the original models and perform well at all CS ratios in a certain range. And we further prove that SDCS outperforms other SSR methods. Finally, we highlight that SDCS can be also applied in other CS-based applications like video CS~\cite{shi2020video} and MRI~\cite{sun2016deep}.


\ifCLASSOPTIONcaptionsoff
  \newpage
\fi

\end{document}